\pdfoutput=1
\documentclass[10pt,journal,compsoc]{IEEEtran}

\usepackage{CJKutf8}
\usepackage{amsmath}

\ifCLASSOPTIONcompsoc
  \usepackage[nocompress]{cite}
\else
  \usepackage{cite}
\fi
 \usepackage{graphicx}

\ifCLASSINFOpdf
\else
\fi
\begin{document}
\begin{CJK*}{UTF8}{gbsn}
%
\title{Self-supervised Image Enhancement Network: Training with Low Light Images Only}
%
%
%
%

\author{Yu~Zhang,
        Xiaoguang~Di,
		Bin~Zhang,
        and~Chunhui~Wang 
\IEEEcompsocitemizethanks{\IEEEcompsocthanksitem Y. Zhang, B. Zhang and C. Wang are with the Department of Electronic Science and technology, Harbin Institute of Technology, Harbin 150001,China.(E-mail: hitzhangyu@qq.com;teamup@yeah.net;wang2352@hit.edu.cn)
\IEEEcompsocthanksitem X. Di is with the Department of Control Science and Engineering,Harbin Institute of Technology, Harbin 150001,China.(E-mail: dixiaoguang@hit.edu.cn)}
\thanks{}}

%
%

\markboth{}%
{Yu \MakeLowercase{\textit{et al.}}: Self-supervised Image Enhancement Network: Training with Low Light Images Only}
%



\IEEEtitleabstractindextext{%
\begin{abstract}
This paper proposes a self-supervised low light image enhancement method based on deep learning. Inspired by information entropy theory and Retinex model, we proposed a maximum entropy based Retinex model. With this model, a very simple network can separate the illumination and reflectance, and the network can be trained with low light images only. We introduce a constraint that the maximum channel of the reflectance conforms to the maximum channel of the low light image and its entropy should be largest in our model to achieve self-supervised learning. Our model is very simple and does not rely on any well-designed data set (even one low light image can complete the training). The network only needs minute-level training to achieve image enhancement. It can be proved through experiments that the proposed method has reached the state-of-the-art in terms of processing speed and effect.
\end{abstract}

\begin{IEEEkeywords}
Low Light Image Enhancement, Self-supervised Learning, Max Entropy, Retinex.
\end{IEEEkeywords}}

\maketitle

\IEEEdisplaynontitleabstractindextext

%
\IEEEpeerreviewmaketitle

\ifCLASSOPTIONcompsoc
\IEEEraisesectionheading{\section{Introduction}\label{sec:introduction}}
\else
\section{Introduction}
\label{sec:introduction}
\fi

%
%
%
%
\IEEEPARstart{R}{ecently}, various algorithms based on deep learning have achieved surprising results in some image processing and computer vision tasks, such as object detection \cite{liu2016ssd}, \cite{ren2015faster}, \cite{redmon2018yolov3},\cite{he2017mask}, image segmentation\cite{he2017mask},\cite{long2015fully},\cite{badrinarayanan2017segnet}, etc. One important reason for the rapid development of deep learning in these tasks is that we can obtain a large number of data sets with clear and unambiguous labels. In these tasks, although the construction of the data set requires some cost, it is still acceptable, also on the Internet, a large number of open source data sets can be found for these tasks to support the training of the network. However, in low-level image processing tasks such as low light image enhancement, image dehazing, and image restoration, etc., it is difficult to obtain a large number of true input/label image pairs.
\par As for low light image enhancement task, in the previous work, some solutions such as  synthesizing low light images \cite{Lore2015LLNet}, using different exposure time images to obtain data \cite{chen2018learning}, and so on, have achieved good visual effects. However, there are still two problems with those methods. One is how to ensure that the pre-trained network can be used for images collected from different devices, different scenes, and different lighting conditions rather than building new training data set. The other is how to determine whether the normal light image used for supervision is the best, there can be lots of normal light images for a low light image. Usually, the builder of the data set gets the normal light images by experience or artificial adjustment, which will cost lots of time and energy and we cannot make sure that the enhanced image can show the information contained in the low light image to the greatest extent with those normal light images.
\par For those two questions, this paper proposes a self-supervised low light image enhancement network based on information entropy theory and Retinex model, and achieves the state-of-the-art in terms of enhancement quality and efficiency. In this paper, the only data we need are the low light images, without any paired or unpaired normal light images. To our knowledge, this is the first fully self-supervised image enhancement method based on deep learning. The proposed method does not rely on a well-designed complex network structure, only with a simple fully convolutional neural network (CNN) as shown in Fig.2 and minute-level training, we can complete low-light image enhancement tasks.
\par There are some image enhancement networks based on Retinex model \cite{wei2018deep},\cite{park2018dual}, but they all require paired data, and then use the assumptions that images captured in different light conditions should have the same reflectance and the illumination map should be smooth to decompose low light images into corresponding reflectance and illumination map. Similar to these works, we also use a network to decompose low light image into reflectance and illumination, but unlike those previous works, we use self-supervised methods to train the network. Only low light images are required (even a single low light image) for training, then we can get the reflectance with good visual effects, and it can be treated as an enhanced image.
\par We think that low light image enhancement task is to display the information contained in low light images in a more intuitive way, rather than creating new information. At the same time, according to the entropy theory, images whose histogram are uniform distribution have the maximum entropy and contain the most information. Based on the above analysis, we propose an assumption that the histogram distribution of the maximum channel of the enhanced image should conform to the histogram distribution of the maximum channel of the low light image after histogram equalization. With this assumption, the loss function can be designed without normal light images, and it can not only retain the authenticity of the enhanced image, but also ensure that the enhanced image has sufficient information. The proposed method does not have any dependence on the way of acquiring low light images, and the training process is completely self-supervised, so the method proposed in this paper has good generalization ability, even if the pre-trained network is not well enough in new environment, retraining or fine-tuning without building paired/unpaired normal light images data set is possible for the network.
Our contributions include:
\begin{itemize}
\item We propose a new maximum entropy based Retinex model, and give its theoretical source.
\item Combined with deep learning, we propose a self-supervised low light image enhancement network, which can complete the training with even one single low light image.
\item The proposed method only requires minute-level training and has a good real-time performance. We verify the enhancement effect and stability of the algorithm through some experiments and objective indexes.
\end{itemize}

 
 



\section{Related Works}

\par Our method mainly comes from histogram equalization, model-based methods, and deep learning based image enhancement methods.
\\
\textbf{Histogram Equalization}

\par In low light image enhancement tasks, Histogram Equalization(HE) is the most simply and wildly used method. It can let the histogram of the enhancement image have a uniform distribution to get the maximum entropy. However, HE cannot avoid the problems of details disappearance (over enhancement, under enhancement), poor color restoration, noise amplification and so on.

\par To solve those problems, various improved algorithms are proposed, such as Adaptive Histogram Equalization(AHE)\cite{pizer1987adaptive} and Contrast-limited Adaptive Histogram Equalization(CLAHE)\cite{pisano1998contrast} for details, Hue-preserving color image enhancement \cite{naik2003hue} for hue preserving, Brightness Bi-Histogram Equalization Method (BBHE) \cite{kim1997contrast}, Dualistic Sub-Image Histogram Equalization Method (DSIHE) \cite{wang1999image} for brightness preserving, etc. In \cite{celik2011contextual} and \cite{lee2013contrast}, the method considering the relationship between adjacent pixels and large gray-level difference is proposed. Although many improved methods have been proposed, there are still many problems in applying histogram equalization directly to image enhancement. 
\begin{figure}[htbp]
\centering
\includegraphics [width=3.5in]{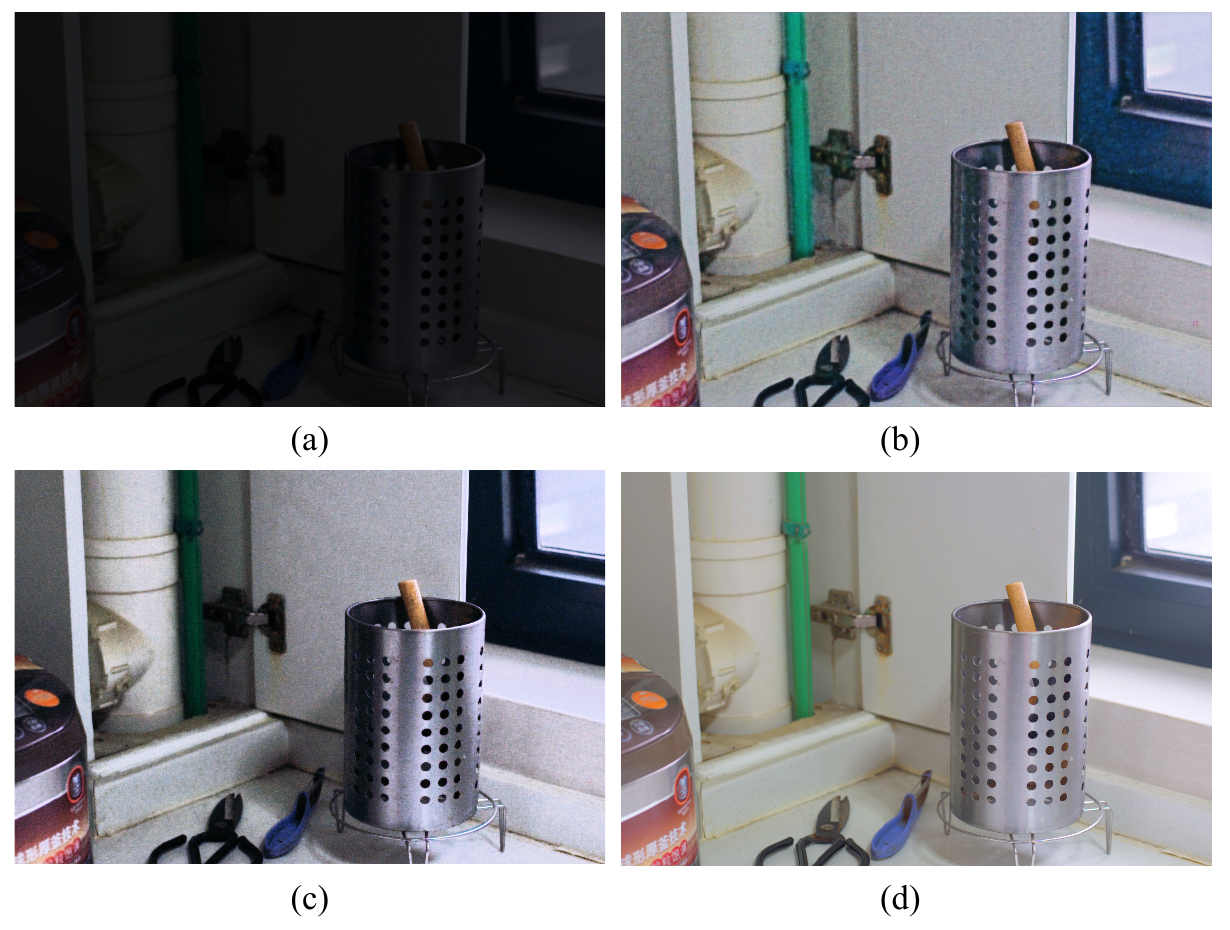}
\caption{The enhancement results by proposed method. (a) Input. (b) The network was trained 200 epochs with low-light images without (a). (c) The network was trained 10000 times with image (a) only. (d) Reference}
\label{fig1}
\end{figure}
\\
\textbf{Model Based Image Enhancement Method}
\par Among the model-based low-light image enhancement methods, there are mainly based on the dehazing model \cite{dong2011fast} and Retinex model\cite{land1977retinex}. The method based on the dehazing model is mainly based on the discovery that the low light image is similar to the haze image after inversion. Dong et al. proposed an enhancement method that performs the dehazing operation after inverting the low light image and then inverts the image back \cite{dong2011fast}. Some studies have extended these works \cite{li2015low}, although these methods have achieved some good effect, they lack corresponding physical model, which limits the application of the method in various scenes. 

\par According to Retinex theory, the collected images can be decomposed into illumination and reflectance, but this is a highly ill-posed problem to obtain them from low light images only. Therefore, other constraints must be introduced: the early researches like single-scale Retinex \cite{jobson1997properties} model and the multi-scale Retinex \cite{jobson1997multiscale} model, only use the constraint that the illumination map is smooth to solve the problem. The captured image is smoothed by using Gaussian filters with one or more scales to obtain illumination, however, the enhanced image often have unreal phenomena such as over-enhancement and whitening. \cite{wang2013naturalness} proposes a Bright-Pass Filter that preserves natural characteristics. \cite{fu2016fusion} performs global illumination adjustment and local contrast enhancement on the initially estimated illumination map. Although they have obtained some good effects, without considering the structural characteristics, information loss is prone to occur in areas with rich details. LIME \cite{guo2016lime} introduces a filter considering the structure characteristic to smooth the illumination map and uses BM3D to denoise the enhanced image. Although those methods are proposed to maintain image details and naturalness, neither before nor after enhancement tasks, the denoising process will still cause blur or loss of details.

\par It is difficult to add some additional priors to those methods based on Retinex model only, which leads to the problems of noise, halo, detail preservation and so on, so in recent years, many algorithms based on the variational Retinex model are proposed. Kimmel et al. \cite{kimmel2003variational} first proposes a variational Retinex model, and uses $L_{2}$ regularization to obtain a smooth illumination map. Fu et al. \cite{fu2013variational} introduces the bright channel prior to the variational Retinex model to suppress the halo effect. Park et al. \cite{park2017low} proposes a weighted $L_{2}$ regularization to constraint reflectance image, which has a slight noise suppression effect. Fu et al. \cite{fu2019hybrid} proposes a $L_{2}$-$L_{p}$ norm to constrain the illumination map and keep more details. Although these variation based methods have achieved good results, it is very time-consuming to process images due to the need of multiple iterations to solve the variational equation. Even with Fast Fourier Transform(FFT), it is difficult to ensure the real-time. 

\par In addition, in those model-based low light image enhancement algorithms, spatial smoothing prior \cite{lagendijk2012iterative} and its improvement are mostly used to constrain the illumination map, and there is no constraint on the contrast information of the reflectance. In this paper, we use the maximum information entropy to constrain the reflectance image, so as to further improve its contrast information.
\\
\textbf{Learning based methods}
\par Learning based methods have achieved good results in some low-level image processing tasks, such as image denoising \cite{zhang2017beyond},\cite{guo2019toward}, super-resolution reconstruction \cite{dong2015image},\cite{li2019feedback},restoration\cite{nah2017deep},\cite{kupyn2018deblurgan}, etc. However, most of the current algorithms based on deep learning are supervised, and it is difficult to obtain both degraded and normal images in those low-level image processing tasks. It is proposed to synthesize low light image data with normal light image for training in some researches. For example, LLNET \cite{lore2017llnet} is the first work to use deep learning to solve image enhancement problem, it proposes to train the networks with synthetically noisy and dark images separately, but it does not consider the natural images characteristic. In \cite{li2018lightennet} and \cite{yang2016enhancement}, gamma transformation is applied to natural image patches to generate low light image patches for training, but they do not consider other degradation of the real collected low light images like noise, color changing, etc. In MSR-net \cite{shen2017msr}, high quality (HQ) data are obtained by artificial selection and Photoshop, and low light images are obtained by processing HQ data with random brightness and contrast reduction and gamma transformation. The data obtained by those methods seems to look like low light images, however, it is difficult to truly reflect the characteristics of low light images, such as noise, overexposed and underexposed areas existed in the same image, etc.

\par In order to solve this problem ,some methods propose to use real low light images for training. In \cite{cai2018learning}, a large multi-exposure image database is established, and the reference images are obtained by combing different exposure images and subjective selection. Retinex-net \cite{wei2018deep} tries to obtain the low/normal light image pairs through adjusting the exposure time, and achieves good enhancement effects, but the exposure time is still artificially determined, and it is difficult to choose the best exposure time to get a reference image. In \cite{chen2018learning}, it introduces a parameter to link two images with different exposure time, and with end-to-end training, it can well deal with the noise problem, but it can only be used for raw images. In \cite{zhang2019kindling}, a light adjustment network is introduced to link paired images with any different exposure time, which solves the problem in acquisition of normal light images. However, in practical applications, if we want to get better images, we may need to choose a hyper-parameter for each low light image. 

\par Although these deep learning based methods have achieved good visual effects in low light image enhancement, they are all based on paired images, and the cost of building training data is so high, and they do not solve the two problems we mentioned before, i.e. how to obtain an optimal reference image and how to ensure the adaptability of the method to new environments or new equipments.

\section{Method}
\subsection{Maximum Entropy Based Retinex model }

\begin{figure}[tp]
\centering
\includegraphics [width=3.5in]{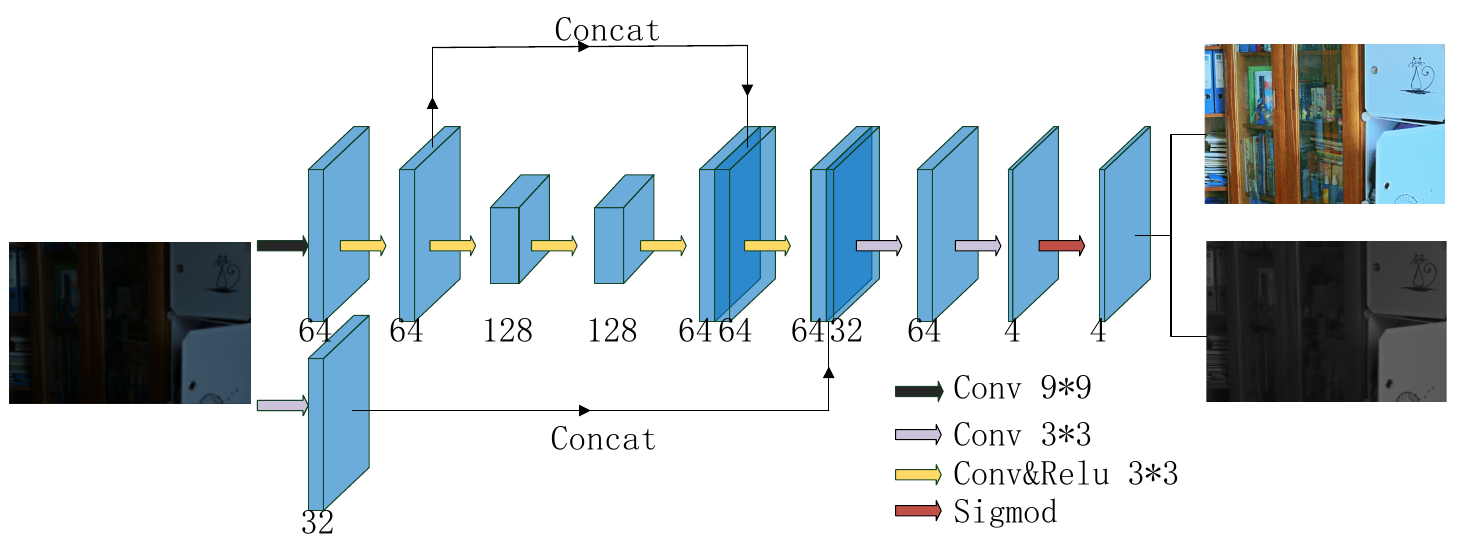}
\caption{Structure of the Self-supervised image enhancement network}
\label{fig2}
\end{figure}

\par Base on Retinex model, an image can be decomposed into reflectance and illumination map as follows: 

\begin{equation}
\label{eqn1}
S=R\circ I
\end{equation}
\noindent where $S$ represents the captured image, $R$ represents the reflectance and $I$ represents the illumination map. This is a highly ill-posed problem, its solution needs additional prior. According to Bayesian formula, the problem can be expressed as follows: 

\begin{equation}
\label{eqn2}
p(R,I\mid S)\propto p(S\mid R,I)p(R)p(I)
\end{equation}

\noindent Where, $p(R,I\mid S)$ is posterior probability, $p(S\mid R,I)$ is the class conditional probability, and $p(R)$ and $p(I)$ are prior probabilities of reflectance and illumination. Existing methods generally add the prior probabilities $p(R)$ and $p(I)$ to find the maximum posterior probability, and estimate the reflectance and illumination.

\par By calculating the negative logarithm of equation (\ref{eqn2}), the problem of image enhancement can be transformed into the form of three distance terms, as can be seen in formula (\ref{eqn3}):

\begin{equation}
\label{eqn3}
\underset{R,S}{min} l_{rcon}+\lambda_{1}l_{R}+\lambda_{2}l_{I}
\end{equation}

\noindent Where, $l_{rcon}$ represents reconstruction loss, $l_{R}$ represents reflectance loss, and $l_{I}$ represents illumination loss. $\lambda_{1}$ and $\lambda_{2}$ are weight parameters.

\par In this paper, we use the $L_{1}$ norm to constrain all the losses, we do not compare the impact of $L_{1}$, $L_{2}$, SSIM and other loss functions on low level image processing tasks, there are some related studies such as\cite{zhao2016loss}. The reconstruction loss $l_{rcon}$ can be expressed as: 

\begin{equation}
\label{eqn4}
l_{rcon}=\left \| S-R\circ I \right \|_{1}
\end{equation}

\par As for the reflectance loss, different from the existing methods only using  $ \left \|\bigtriangleup R \right \|_{1}$ \cite{park2017low},\cite{fu2015probabilistic}, we propose a new distance measurement method for reflectance loss based on the following reasons:

\begin{itemize}
\item For image enhancement task, the processed image should have enough information
\item The processed image shall conform to the original image information
\item Histogram equalization can greatly improve the information entropy of image
\end{itemize}

\par Based on the above considerations, we propose equation (\ref{eqn5}) as the loss of reflectance image, which also uses $L_{1}$ loss:

\begin{equation}
\label{eqn5}
l_{R}=\left \| \underset{c\in{R,G,B}}{max} R^{c} - F(\underset{c\in{R,G,B}}{max}S^{c}) \right \|_{1} + \lambda \left \|\bigtriangleup R \right \|_{1}
\end{equation}

\noindent Where, $F(X)$ means the histogram equalization operator to image X. $\lambda$ is weight parameters. This loss function means that maximum channel of the reflectance should conform to the maximum channel of the low light image and has the maximum entropy. There are three main reasons why we choose the maximum channel to constrain. Firstly, for a low light image, the maximum channel has the greatest impact on its visual effect. Secondly, if other channels are selected, there is no doubt that saturation will occur according to the prior that the maximum channel must be greater than the other two channels. Thirdly, if we choose one of the color channel, such as R, G or B channel, it is obviously not in line with the natural image. 

\par For the illumination loss, we adopt the structure-aware smoothness loss proposed in \cite{wei2018deep}:

\begin{equation}
\label{eqn6}
l_{I}=\left \| \bigtriangleup I\circ exp\left ( -\lambda_{3}\bigtriangleup R  \right ) \right \|_{1}
\end{equation}
\par It is proposed that equation (\ref{eqn6}) can make the illumination loss aware of the image structure in \cite{wei2018deep}. And the loss means that the original TV function  $\left \| \bigtriangleup I \right \|_{1}$ is weighted with the gradient of reflectance.

\par From the equation (\ref{eqn3}) to equation (\ref{eqn6}), we get the maximum entropy based Retinex model, as can be seen in equation (\ref{eqn7}):

\setlength{\arraycolsep}{0.0em}
\begin{eqnarray}
\label{eqn7}
Z&{}={}&\left \| S-R\circ I \right \|_{1}+\lambda_{1} \left \| \underset{c\in{R,G,B}}{max} R^{c} - F(\underset{c\in{R,G,B}}{max}S^{c}) \right \|_{1} \nonumber\\
&&{+}\:\lambda_{2} \left \| \bigtriangleup I\circ \lambda exp\left ( -\lambda_{3}\bigtriangleup R  \right ) \right \|_{1} \nonumber\\
&&{+}\:\lambda_{4}\left \|\bigtriangleup R \right \|_{1}
\end{eqnarray}
\setlength{\arraycolsep}{5pt}

\par Variational methods or FFT are generally used to solve equation (\ref{eqn7}) with $L_{2}$ loss, however, they both need multiple iterations which will bring time consumption problems, and with more constraints, the solution will be more complicated.  In order to enhance the image in real time, we propose a solution based on deep learning. The network uses equation (\ref{eqn7}) as the loss function. We can find that in equation (\ref{eqn7}), there is only low light images, so the network can be trained through a self-supervised way. 
\par The values of $\lambda_{1}$,$\lambda_{2}$,$\lambda_{3}$,$\lambda_{4}$ are $0.1$, $0.1$, $10$ and $0.01$ in this paper. The influence of values of $\lambda_{1}$ and $\lambda_{2}$ is not so obvious in visual effect that we just choose 0.1, the value of $\lambda_{3}$ comes from \cite{wei2018deep}. As for $\lambda_{4}$, in our experiments, we found that it can be used to control the noise. When its value increases, the noise decreases, and at the same time, the image will be more blurry. Through some experiments, we choose 0.01 for $\lambda_{4}$ and if $\lambda_{4}=0.1$, the enhanced image will appear obvious blur. 

\subsection{Self-supervised Network Based Solution}
\par If we use the variational methods or FFT to solve the model proposed from equation (\ref{eqn3}) to equation (\ref{eqn7}), then it means that we need to carry out the same iterative processing for each low light image, which will not only bring time-consuming problems, but also the iteration times for each low light image may be uncertain, which is almost a disaster in many real applications. At the same time, this kind of solution can not take advantage of big data,  the previous data processing can do nothing helpful to the new data processing. 
\par In the previous deep learning based researches, due to the lack of models that can support self-supervised training, only the paired or unpaired low/normal light images collected in advance can be used to complete the network training. However, the data collected in advance can not contain all the real low light situations, such as different environments, devices, or degradation problems, etc., which also limits the application scope of the pre-trained network.  After all, it is impossible to build the data set when we are using them.
\par However, based on the model proposed from equation (\ref{eqn3}) to equation (\ref{eqn7}), we can achieve the self-supervised training, which means that we can build the data set online and avoid the problem of applicability. And compared with the supervised learning whose supervisor is selected by artificial method, the model based on maximum entropy can ensure that the enhanced image has enough information entropy. 
\par We only need a very simple CNN structure to achieve the decomposition of the illumination and reflectance. The specific structure of the CNN we finally adopted is shown in Fig.2. The input of the network is low light image and its maximum channel, after some convolution and concat layers, reflectance and illumination can be gotten with a sigmod layer. Table \ref{tab:structure} is the specific information of each layer of the network.
\par In fact, we have experimented with different network structures, and the stacking of convolutional layers and a sigmod layer can also produce acceptable results. However, if we add some concat layers, the enhancement results will become clearer. It can be seen that we use down-sampling and up-sampling in the network, its prime function is to reduce the noise. In some experiments, we find that adding the down-sampling layer will make the image blur, however, it will reduce the noise too.   

\begin{table}
	\centering
	\caption{Self-supervised image enhancement network}
	\resizebox{0.48\textwidth}{!}{
		\begin{tabular}{cccccc}
			\hline
			Inputs & Operator& Kernel  &Output Channels & Stride & Output Name \\ 
			\hline
			RGB\&max{R,G,B} & Conv\&ReLU   & $3\times3$ & 32  & 1 				& Conv0 \\
			RGB\&max{R,G,B} & Conv         & $9\times9$ & 64  & 1 				& Conv \\
			Conv            & Conv\&ReLU   & $3\times3$ & 64  & 1 				& Conv1 \\
			Conv1           & Conv\&ReLU   & $3\times3$ & 128 & 2$\downarrow$   & Conv2 \\
			Conv2           & Conv\&ReLU   & $3\times3$ & 128 & 1 				& Conv3 \\
			Conv3           & Conv\&ReLU   & $3\times3$ & 64  & 2$\uparrow$     & Conv4 \\
			Conv4\&Conv1    & Concat       & -          & 128 & - 				& Conv5 \\
			Conv5		   & Conv\&ReLU   & $3\times3$ & 64  & 1 				& Conv6 \\
			Conv6\&Conv0    & Concat       & -          & 96  & - 				& Conv7 \\
			Conv7           & Conv         & $3\times3$ & 64  & 1 				& Conv8 \\
			Conv8 		   & Conv\&ReLU   & $3\times3$ & 4   & 1 				& Conv9 \\
			Conv9           & Sigmoid      & -          & 4   & -				& R\&I \\
			\hline
		\end{tabular}
	}
	\label{tab:structure}
\end{table}

\section{Experiment}
\par We use the LOL database \cite{wei2018deep} which contains 500 low/normal light image pairs, 485 of which are used for training and images size are $400*600$. Note that during the training process, we only use natural low light images and do not use synthetic data and normal light images. During the training process, our batch size is set to 16 and the patch size is set to 48 * 48. We use Adam stochastic optimization \cite{kingma2014adam} to train the network and the update rate is set to 0.001. The training and testing of the network are completed on a Nvidia GTX 2080Ti GPU and Inter Core i9-9900K CPU, and the code is based on the tensorflow framework.
\par In section 4.1, we introduce some objective evaluation indexes. In section 4.2, we measure the influence of the training times on loss and evaluation indexes. In section 4.3, we measure the stability of the algorithm through repeated experiments. In section 4.4, we compare our algorithm with some existing methods. In section 4.5, we give some enhancement results when the network is trained with one single low light image.

\subsection{Evaluation Indexes}
\par There are many indexes with or without reference that can be used to evaluate the quality of the enhanced image. However, the constrain we use in this paper does not conform to the natural image characteristics, so it is difficult for us to evaluate the enhanced image accurately with those existing evaluation indexes. In this paper, we use gray entropy (GE), color entropy (CE, color entropy is the sum of entropy of {R,G,B} channels), gray mean illumination (GMI), gray mean gradient (GMG), LOE\cite{wang2013naturalness}, NIQE\cite{mittal2012making}, PSNR, SSIM to evaluate the enhanced image. It should be noted that these indexes can only reflect the image quality in some aspects, which are not completely consistent with the evaluation results given by the human visual system. The LOE$_{low}$ and LOE$_{high}$ are calculated with low and high light images respectively. 

\subsection{The Influence of Training Times}
\par We use 485 low light images in the LOL dataset for training, and 15 for testing. Considering that our method is self-supervised, it lacks an absolute reference, and some parameters and constraints in our loss function come from the individual experience. We cannot determine whether our training has reached the best through the change of the loss. So we train the network for 1,000 epochs, and process the testing data every 20 training epochs and use those indexes to evaluate the training results of the network. 
\par Fig.3 and Fig.4 show the change of loss and indexes with the increase of training times. It can be seen that the loss falling fast at the beginning. On our GPU, it takes less than 0.65s to train one epoch. Fig.5 shows enhancement results of low light images in the testing data with different training times. We only selected the results of the first 200 epochs to display. It can be seen that as the training goes, some indexes which can reflect the image clarity such as entropy and gradient increase, however, the gap between the enhanced images and reference images is also growing. That is caused by noise, although the image becomes more and more clear as the training goes, at the same time, the noise keep increasing too. In order to keep balance between clarity and noise, we just stop training after 200 epochs. And in our experiment, if the training epochs keep increasing more than about 1000 epochs, there will be artifacts in some testing images like \cite{chen2018learning}. Early stopping is a reasonable method to avoid the noise and artifacts.
 
\begin{figure}[htbp]
\centering
\includegraphics [width=3.5in]{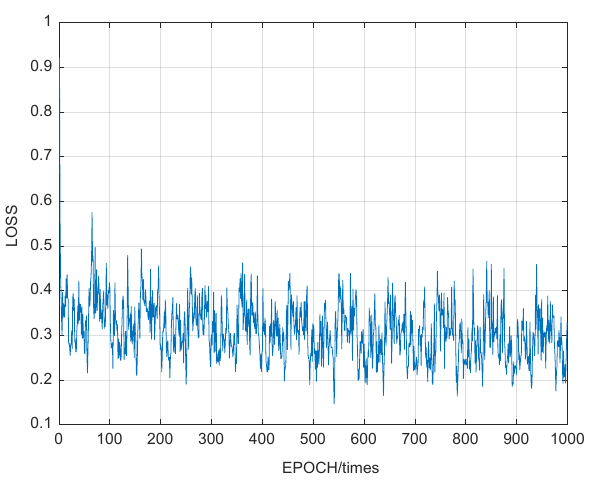}
\caption{The training loss with 1000 epochs}
\label{fig3}
\end{figure}

\begin{figure*}[htbp]
\centering
\includegraphics [width=\linewidth]{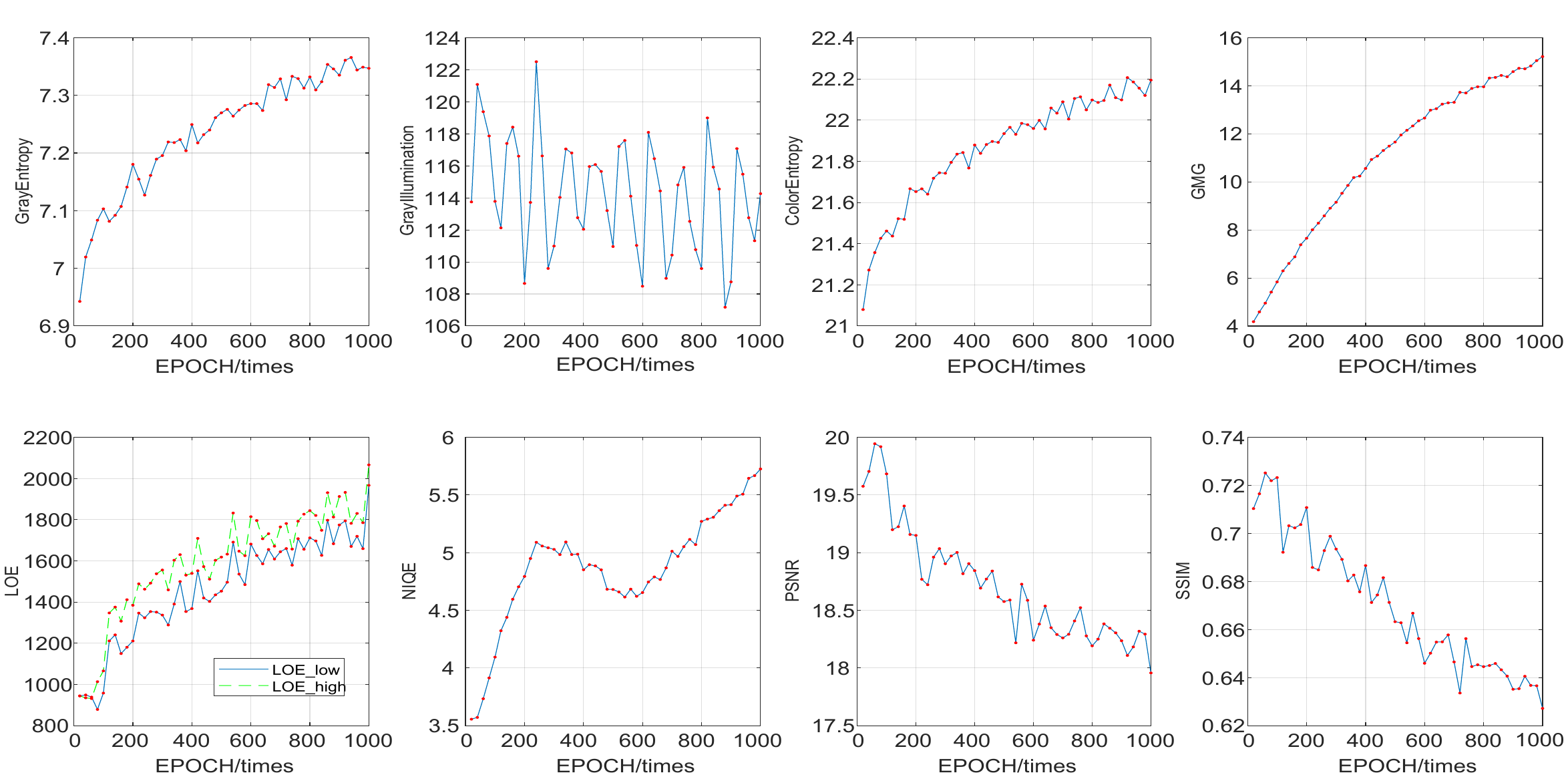}
\caption{Evaluation indexes on the testing data by different training epochs}
\label{fig4}
\end{figure*}

\begin{figure*}[htbp]
\centering
\includegraphics [width=\linewidth]{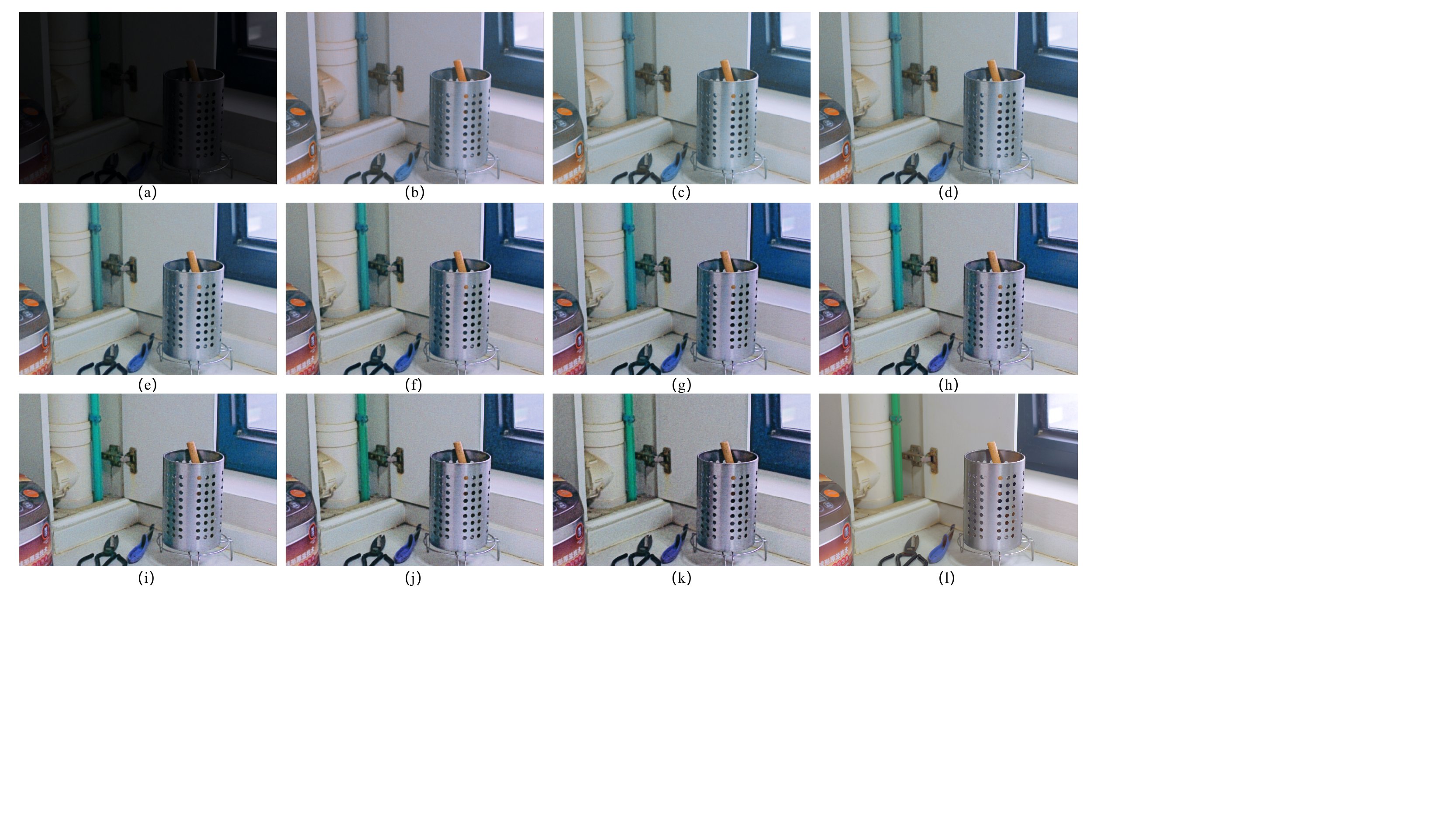}
\caption{The enhancement results by different training epochs (a) Original. (b) Epoch = 20
(c) Epoch = 40. (d) Epoch = 60. (e) Epoch =80. (f) Epoch =100. (g) Epoch =120. (h) Epoch =140. (i) Epoch =160. (j) Epoch =180. (k) Epoch =200. (l) Reference.}
\label{fig5}
\end{figure*}

\subsection{Repeated Learning Stability}
\par Due to the characteristics of learning based methods, in most of cases, we cannot reproduce the optimal results. So we repeat the experiments many times to evaluate the repeatability of the method. In every experiment, we train the network with 200 epochs, and evaluate the network on testing data through indexes mentioned in section 4.1. Fig.6 and Fig.7 show the evaluation indexes and some enhancement results in different experiments respectively. It can be seen that there are large fluctuations in some indexes, like LOE、GMG and NIQE, however, the changings of enhancement results are not so obvious in most experiments. In the fifth experiment, the color of enhanced images are lighter than others. We think that the differences of enhancement results may come from the $L_{1}$ loss functions and the difference among the training data in every experiment.
\par In every experiment, the only difference is training patch, which seems to have an impact on training results. Those training patches are randomly selected and cropped, and considering the training times and the large size difference between images and patches, they are only a small part of training images. At the same time, we use the $L_{1}$ loss for training, compared with $L_{2}$ loss, $L_{1}$ loss may have multiple solutions, and its solutions will be highly affected by training data. When training data changes, the results may also change greatly. However, from visual effects, the method proposed in this paper is relatively stable. 
\begin{figure*}[htbp]
\centering
\includegraphics [width=\linewidth]{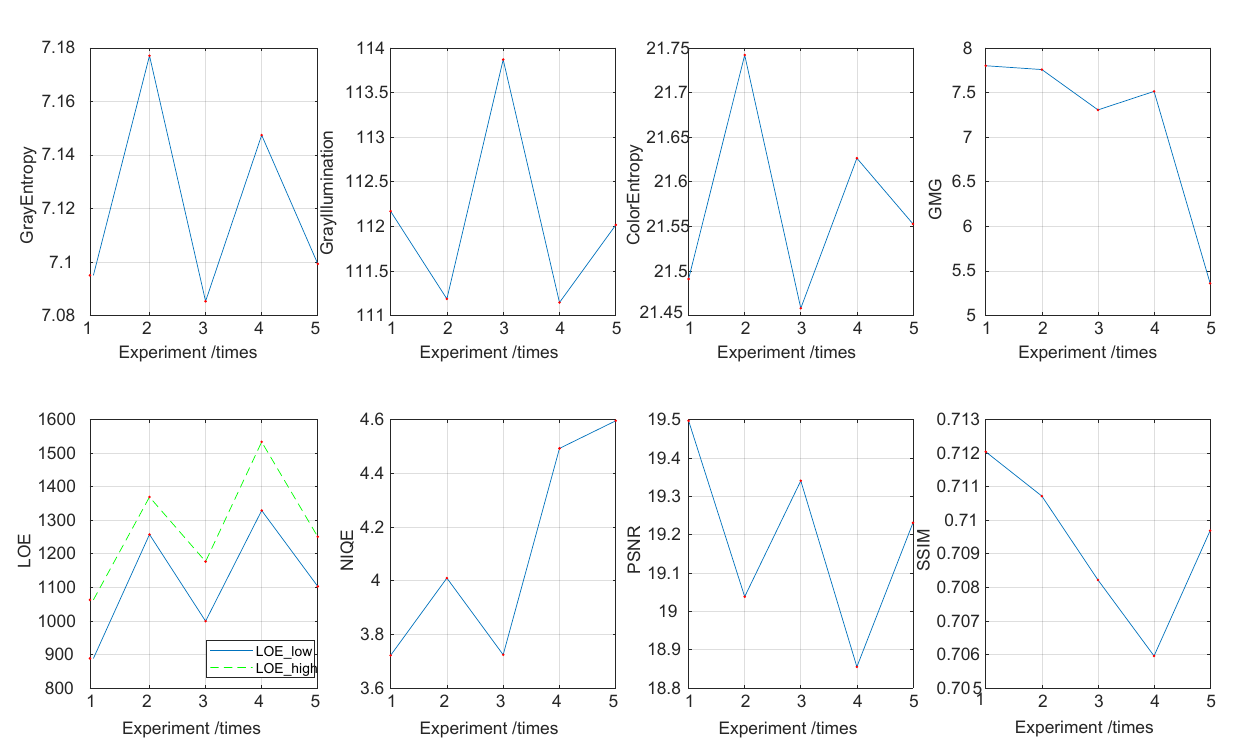}
\caption{Index changes in repeated experiments, 200 epochs training in each experiments 
}
\label{fig6}
\end{figure*}

\begin{figure*}[htbp]
\centering
\includegraphics [width=\linewidth]{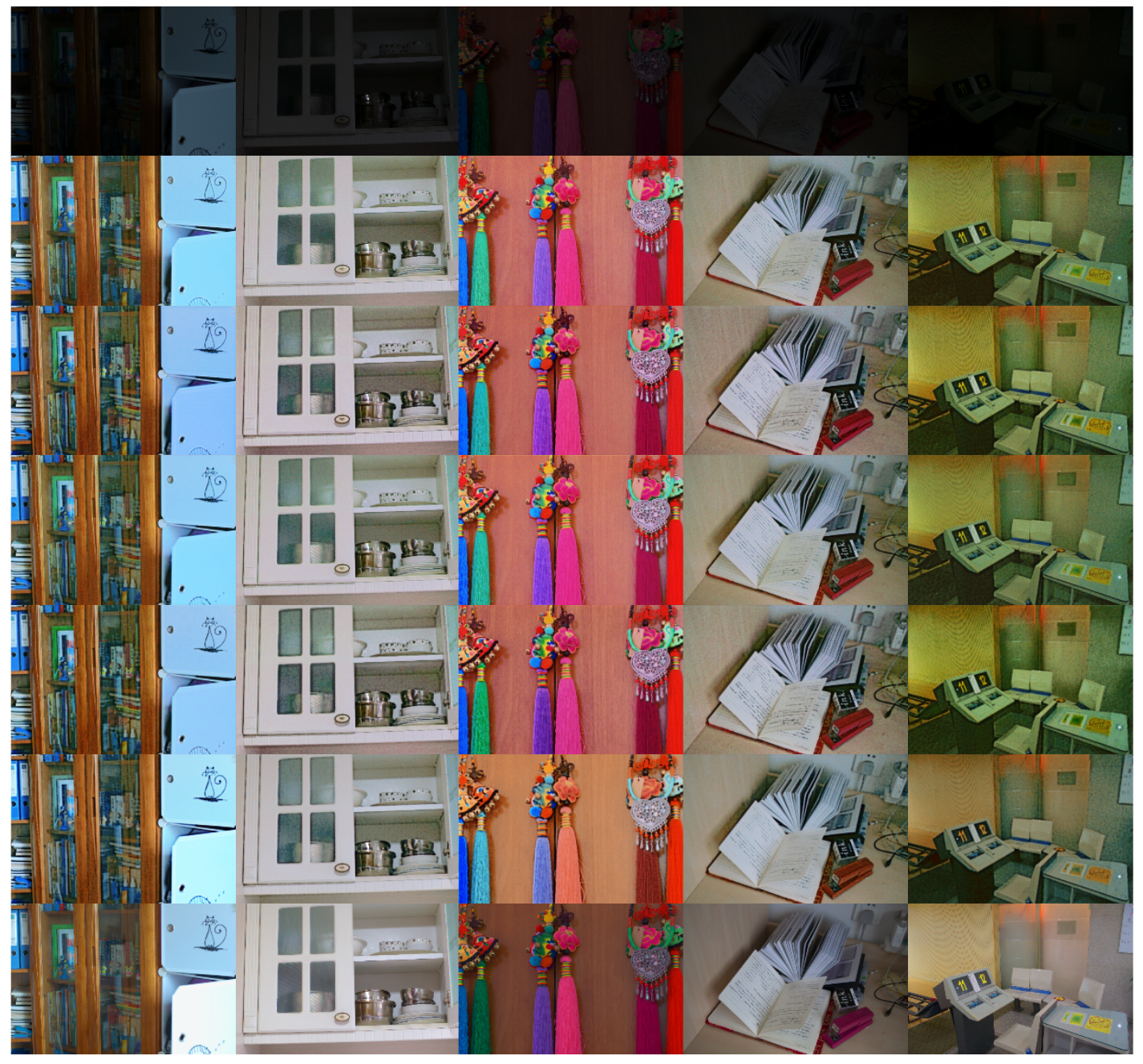}
\caption{The first row is the original low light image. The 2-6 rows are different experiments. The last row is the reference normal light image 
}
\label{fig7}
\end{figure*}

\begin{table*}[htbp]
	\centering
	\begin{tabular}{c|cccccccccc}
		\hline
		Metrics &GE  &GMI  &CE  &GMG  &LOE$_{low}$  &LOE$_{high}$  &NIQE  &PSNR  &SSIM  &Time \\ 
		\hline
		HE      			&\textbf{7.785}    &105.3	    &\textbf{23.21}	&17.12	&\textbf{505.3}	&\textbf{898.6}	&5.201	&15.81	&0.5607	&0.0158  \\
		MSR     			&6.947     &134.7	&17.61	&16.93	&540.3	&950.1	&8.008	&16.69	&0.5262	&0.0911 \\
		NPE   			&7.070    &96.67	    &20.787	&20.38	&1607.7	&1867.8	&9.135	&16.97	&0.5894	&4.71 \\
		MF			   	&6.937     &85.99	&20.16	&17.47	&840.9	&1197.7	&9.713	&16.97	&0.6049	&0.128 \\
		SRIE   			&6.296     &50.33	&18.62	&9.380	&952.0	&1291.4	&7.535	&11.86	&0.4978	&3.45 \\
		LIME    			&7.564    &114.9	    &21.72	&\textbf{23.46}	&1303.5	&1543.7	&9.127	&16.76	&0.5644	&0.211\\
		Gladnet     		&7.116     &119.0	&21.52	&9.918	&902.5	&1205.5	&6.797	&\textbf{19.72}	&0.7035	&0.0212 \\
		Retinex-Net   	&6.835     &110.2	&21.13	&24.00	&1990.1	&1988.8	&9.730	&16.77	&0.5594	&0.0207 \\
		$L_{2}-L_{p}$ 	&6.419     &51.92	&19.16	&8.704	&933.4	&1250.4	&6.234	&12.15	&0.5103	&5.58 \\
		Ours   			&7.180    & 108.7	&21.65	&7.653	&1210.8	&1384.1	&4.793	&19.15	&\textbf{0.7108}	&\textbf{0.0145} \\
		Ours-single	   	&7.030     &88.08	&20.94	&10.27	&529.0	&1028.2	&\textbf{4.422}	&14.18	&0.5169	&\textbf{0.0145} \\
		Reference   	    &7.040     &115.5  	&21.31	&6.910	&921.9	&-	    &4.253	&-	    &1	&- \\
		\hline
	\end{tabular}
	\vspace{10pt}
	\caption{Quantitative comparison on LOL dataset in terms of PSNR, SSIM, LOE$_{low}$, LOE$_{high}$, and NIQE. The best results are highlighted in bold.}
	\label{tab:LOL}
\end{table*}

\begin{figure*}[htbp]
\centering
\includegraphics [width=\linewidth]{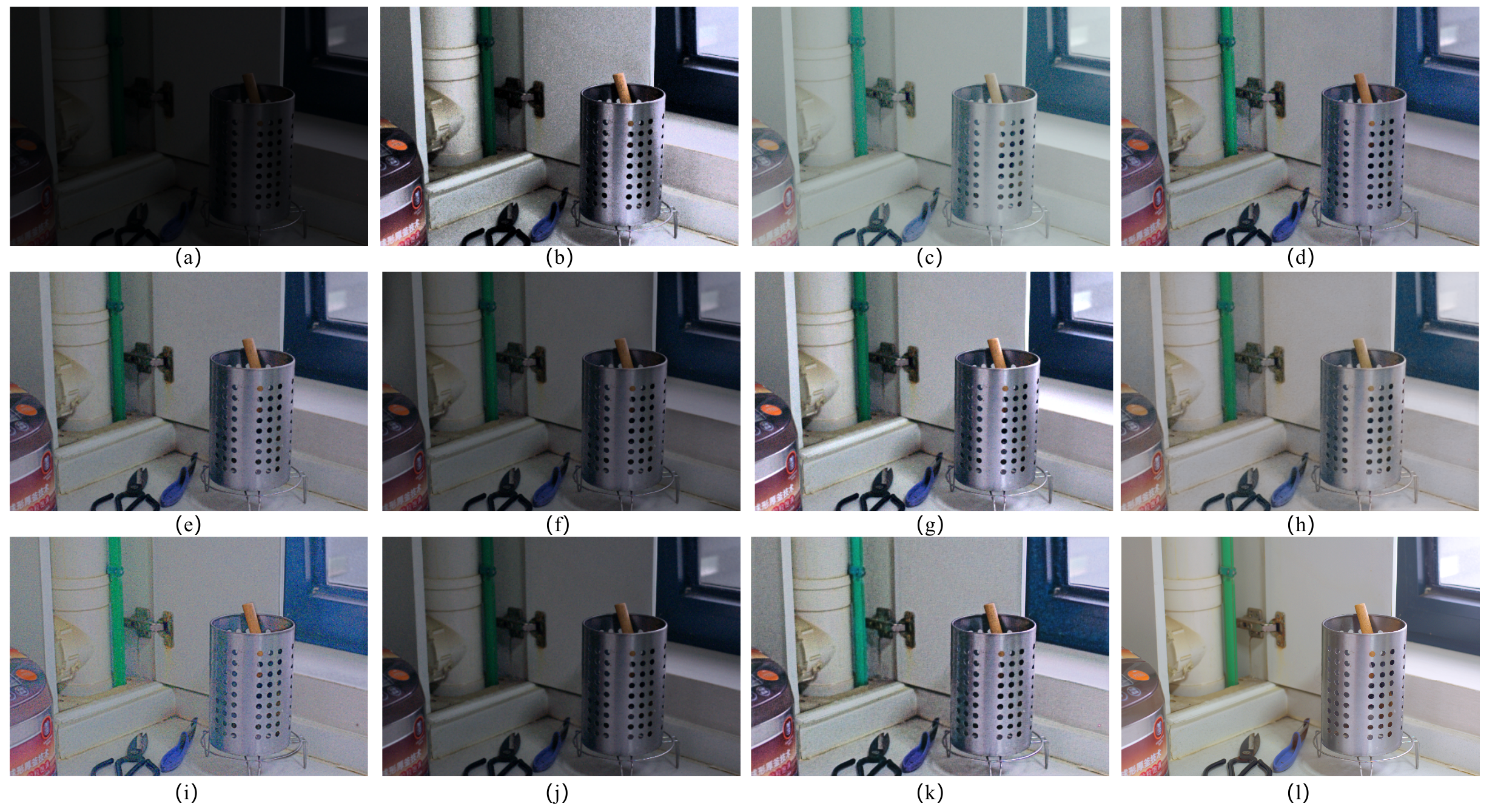}
\caption{The enhancement results by different methods (a) Original. (b) HE. (c) MSR (d) NPE. (e) MF. (f) SRIE. (g) LIME. (h) Gladnet. (i) Retinex-Net. (j) $L_{2}$-$L_{p}$. (k) Ours. (l) Reference.
}
\label{fig8}
\end{figure*}

\begin{figure*}[htbp]
\centering
\includegraphics [width=\linewidth]{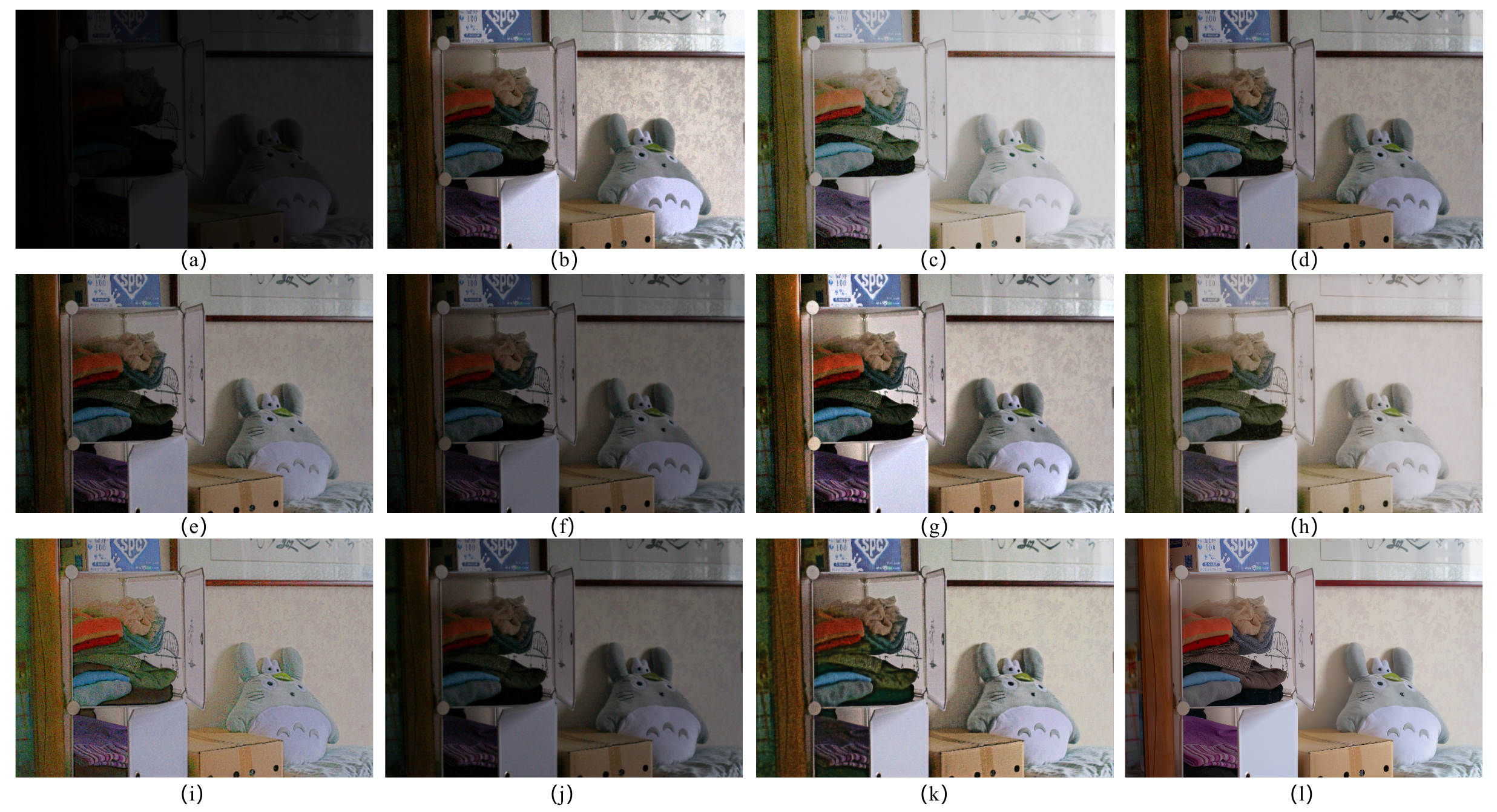}
\caption{The enhancement results by different methods (a) Original. (b) HE. (c) MSR (d) NPE. (e) MF. (f) SRIE. (g) LIME. (h) Gladnet. (i) Retinex-Net. (j) $L_{2}$-$L_{p}$. (k) Ours. (l) Reference.
}
\label{fig9}
\end{figure*}

\begin{figure*}[htbp]
\centering
\includegraphics [width=\linewidth]{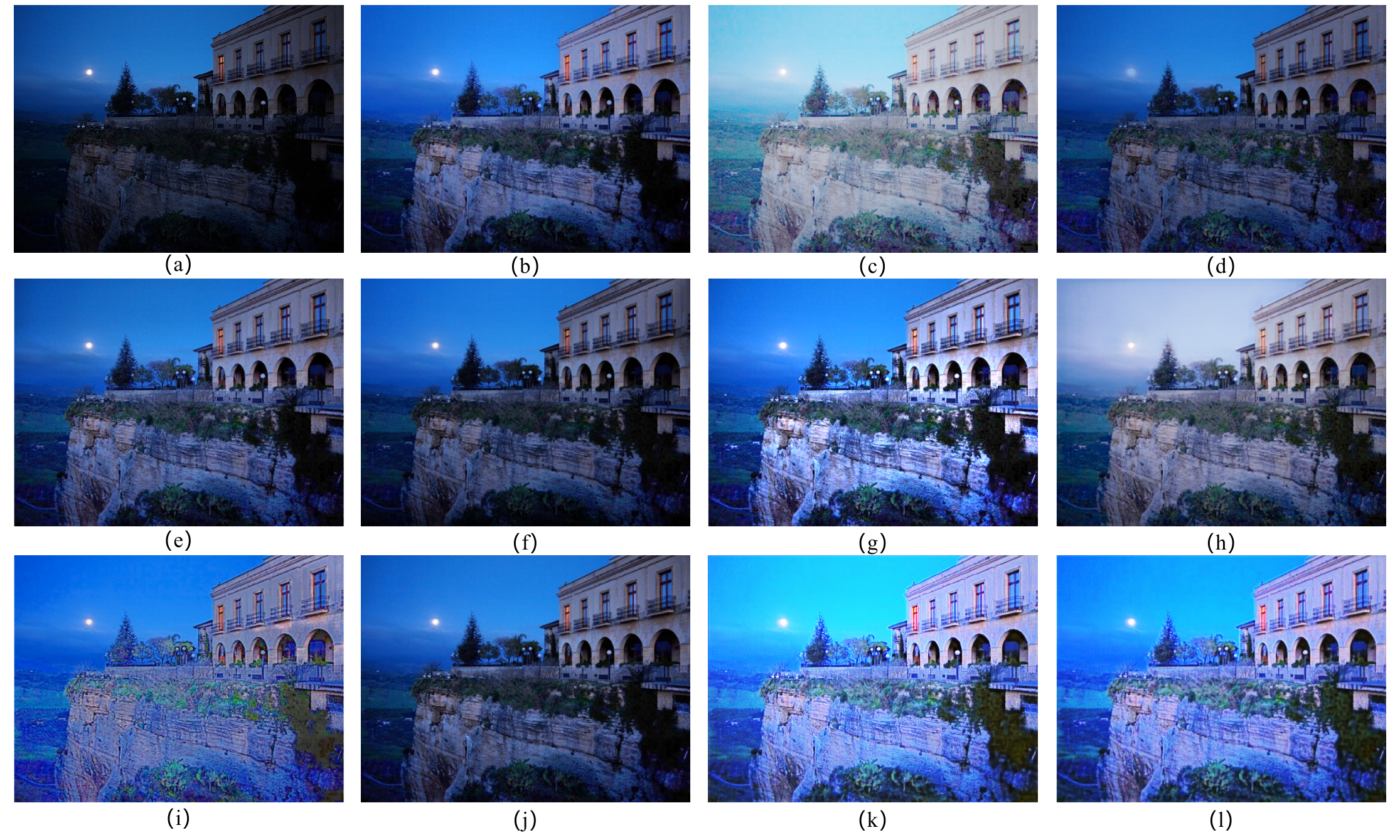}
\caption{The enhancement results by different methods (a) Original. (b) HE. (c) MSR (d) NPE. (e) MF. (f) SRIE. (g) LIME. (h) Gladnet. (i) Retinex-Net. (j) $L_{2}$-$L_{p}$. (k) Ours:Epoch=120. (l) Ours:Epoch=200.
}
\label{fig10}
\end{figure*}

\subsection{Comparisons with Existing Algorithms}
\par We have also compared our algorithm with some existing classic and state-of-the-art methods, including HE, MSR\cite{jobson1997multiscale}, LIME\cite{guo2016lime}, MF\cite{fu2016fusion}, NPE\cite{wang2013naturalness},SRIE\cite{fu2016weighted}, Gladnet\cite{wang2018gladnet}, Retinex-Net \cite{wei2018deep}, $L_{2}$-$L_{p}$\cite{fu2019hybrid}.
\par The 15 images from the LOL dataset are used for testing to get the objective indexes, and Fig.8 and Fig.9 show some enhancement results by different methods, and Table 2 displays the objective indexes and time consumption of those methods. The low light image of Fig.10 is from the LIME \cite{guo2016lime} dataset. All the results of our method come from a randomly selected experiment. 
\par SSIM is generally used to measure the structural similarity between two images. NIQE is a non reference image quality evaluation method. These two indexes can show that our method has good structural similarity and image quality after processing.
\par In CE, GE and GMG, we can see that our method is lower than some methods. Although the larger these indexes are, the more abundant information these images have and the clearer these images are, we also need to consider that these indexes will be highly affected by noise. It can be seen that compared with most of the methods, our method is closer to the reference image in these indexes.
\par In LOE$_{low}$ and LOE$_{high}$, our method does not perform well. This is probably because that the overexposure area and underexposure area may exist in low light images or the reference images at the same time, and the model proposed in this paper can avoid this problem to a certain extent, which leads to our poor performance in these two indexes. As shown in the lower left corner of Fig.7, there are overexposure areas in the reference image. If we use such reference image for training, we can not promise that the training results will not be over exposed. It can also explain that it is difficult to obtain the optimal reference image by adjusting the exposure time.            
\par In PSNR, although our method is lower than Gladnet\cite{wang2018gladnet}, we need to pay attention to that our method does not consider the reference image in training, so it is difficult to ensure the similarity of the enhanced image and the reference image in brightness, which has a certain impact on PSNR. 
\par Although our method can not achieve the best results in all indexes, in terms of visual effect and some important indexes, our method achieves the state-of-the-art. Compared with HE method, our method has a slight denoising effect and can better maintain the structure information and color information. As shown in Table 2, the PSNR and SSIM of our method is higher than the HE method. HE method is not suitable for heavy noise environment and it can be seen in Fig.8 and Fig.9, there are still dark areas in the image after directly using HE method, which is caused by the theory of histogram equalization itself.   Compared with the model-based methods, our method cost less running time. It can be seen in Table 2, our method is more than 6x faster than MSR\cite{jobson1997multiscale} which has the least time among model-based method. Compared with the learning based methods, our method does not need to build data sets carefully, which can save a lot of time and energy and has better applicability for new environment and equipment.

\subsection{Training with Single Low Light Image}
\par At the same time, in order to further evaluate the performance of the method proposed in this paper, we do an experiment that train the network with single low light image. The training image is one of the test data of the LOL dataset and the test data is all the 15 test data of LOL dataset. In Fig.12, we use image 12-(a) only to train the network, and 12-(b) to 12-(k) are the enhancement results of image 12-(a) by different training epochs. Fig.11 displays the evaluation indexes on the test data by different training epochs. Fig.13 shows the enhancement results on some LOL test data and other low light images. Table 2 shows the indexes on the test data by 10000 training epochs. Through Fig.11 to Fig.13, we can see that our method can be applied to new environments quickly, even if we only have one image of the new environment.
\par It can be seen that in terms of visual effect and some indexes, the result of single image training is worse than that of multiple images training. However in our experiment, with the increase of training times, single image training does not produce artifacts. That can prove that the artifacts are not produced by the model proposed in this paper. As we train the image with single low light image, there is no need for the network to fit the histogram equalization stretch, that is an main reason why there are no artifacts in single low light training. It is perhaps difficult to fit the histogram equalization stretch when the network is trained without considering the whole image information in multiple images training. We think that if we want to avoid the artifacts in multiple images training, we have to make the network deeper or considering the whole image information, however, that will also increase time consumption.

\begin{figure*}[htbp]
\centering
\includegraphics [width=\linewidth]{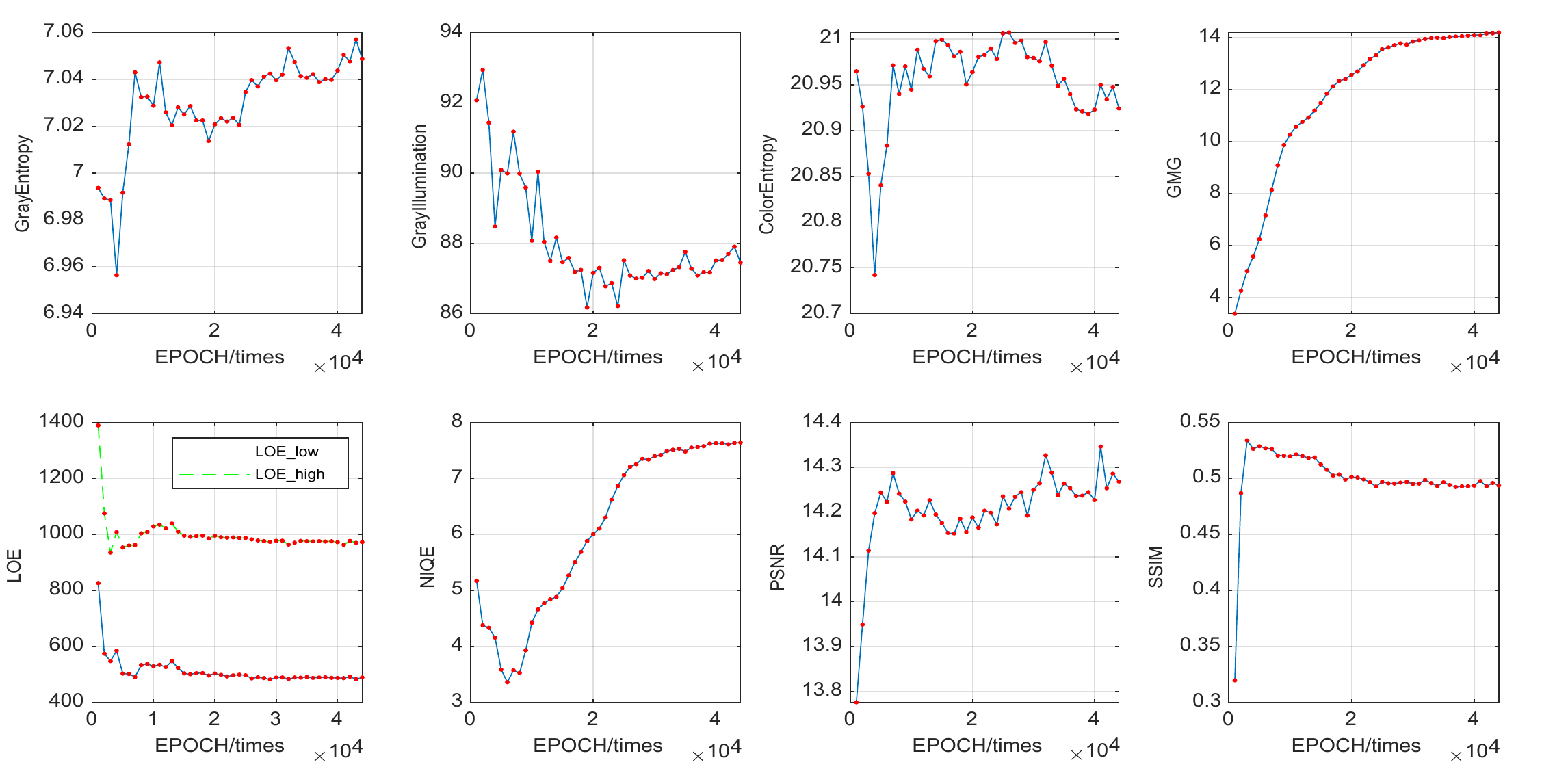}
\caption{Evaluation indexes on the testing data by different training epochs. The network is training with single low light image.
}
\label{fig11}
\end{figure*}

\begin{figure*}[hp]
\centering
\includegraphics [width=\linewidth]{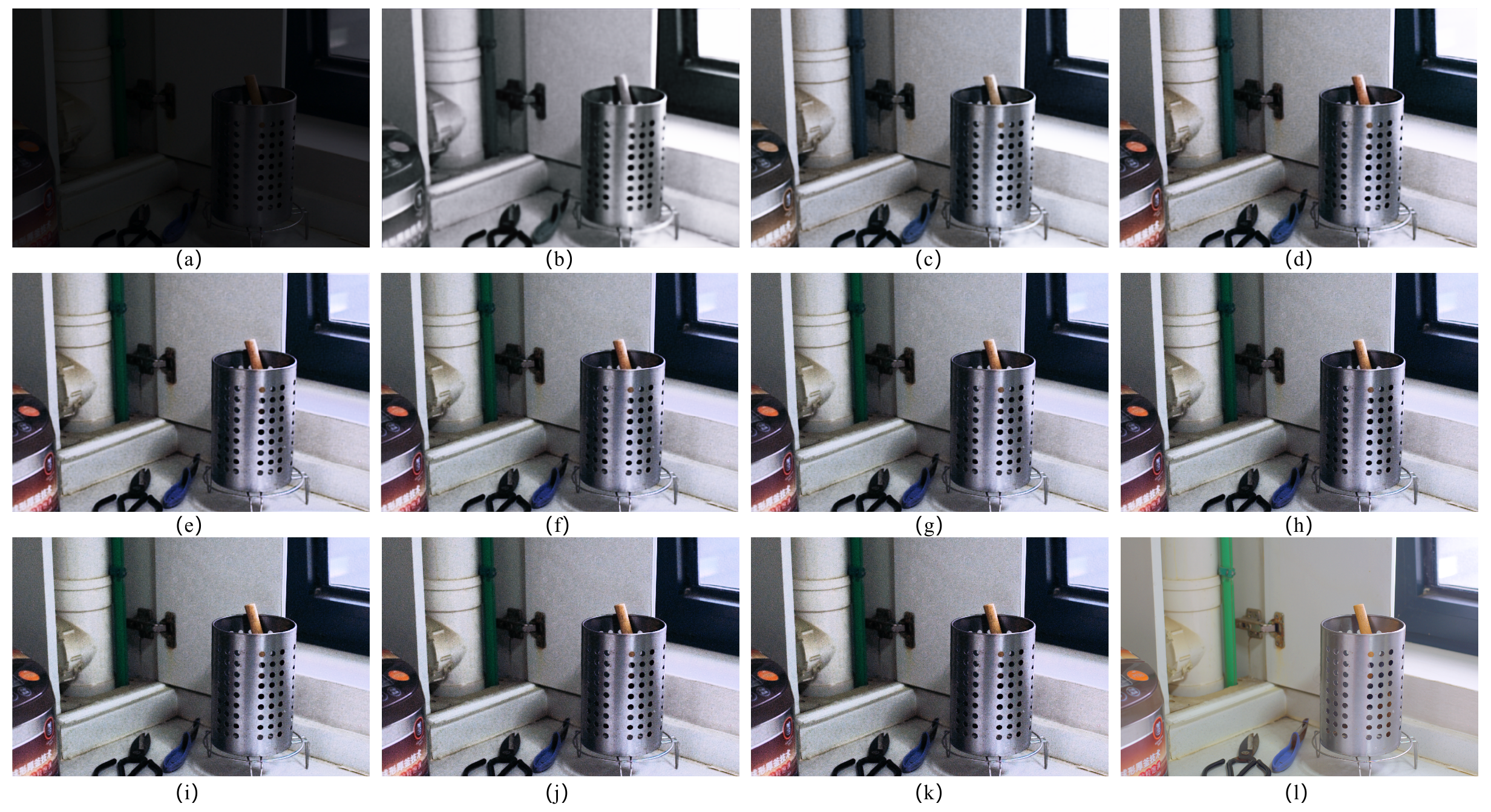}
\caption{The enhancement results with different training times and the network is trained with image (a) only. (a) Original. (b)Epoch=1000. (c) Epoch=2000. (d) Epoch=3000. (e) Epoch=4000. (f) Epoch=5000. (g) Epoch=6000. (h) Epoch=7000. (i) Epoch=8000. (j) Epoch=9000. (k) Epoch=10000. (l) Reference.
}
\label{fig12}
\end{figure*}

\begin{figure*}[hp]
\centering
\includegraphics [width=\linewidth]{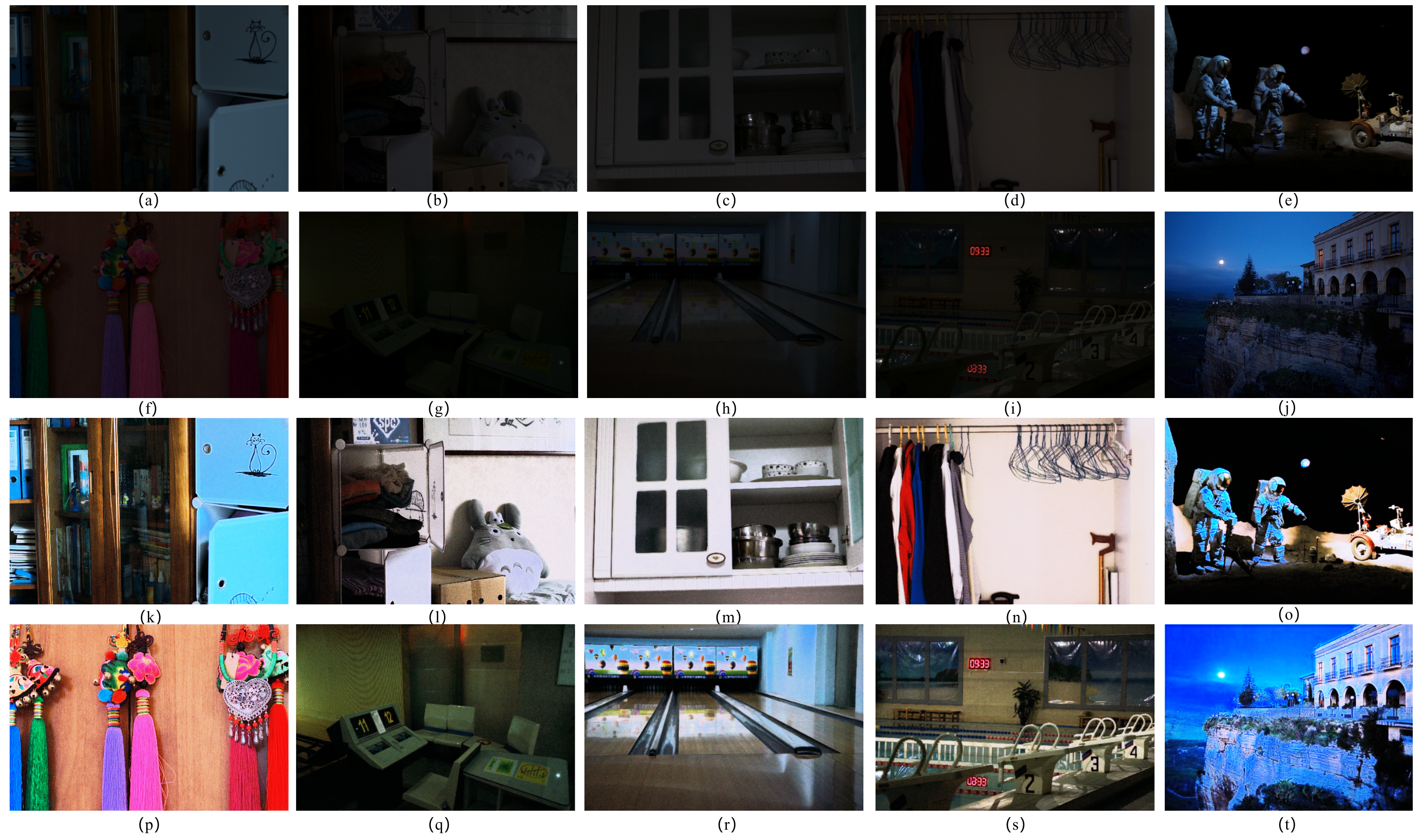}
\caption{The enhancement results and the network is trained by 10000 epochs with image 12-(a) only. (a)-(h) are original low
light images. (i)-(p) are enhancement results.
}
\label{fig13}
\end{figure*}

\section{Conclusion}
\par In this paper, we propose a maximum entropy based Retinex model and a self-supervised image enhancement network. The network can be trained with low light images only and can slightly reduce the noise during enhancement. By testing on real low light images, it shows that, with short time training, the network can produce a well visual effect and has a good real-time performance. It should be noted that our method is self-supervised, so it can adapt to new environments and devices, also, the enhanced image may be different from the real data and look more like the night one in color. The future work will focus on the color restoration, noise and artifact suppression, better detail keeping and so on, we think those can be achieved through Generative Adversarial Networks or new constrains.


%





\ifCLASSOPTIONcaptionsoff
  \newpage
\fi



%

\bibliographystyle{IEEEtran}
\bibliography{IEEEabrv,mybibfile}



%

\begin{IEEEbiography}[{\includegraphics[width=1in,height=1.25in,clip,keepaspectratio]{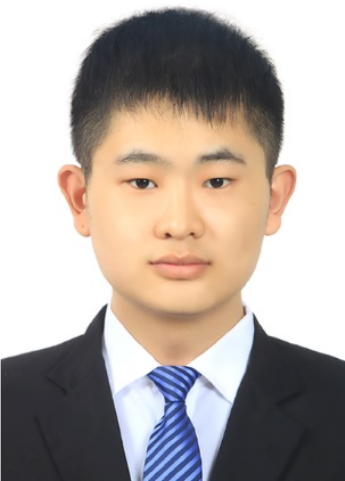}}]{Yu Zhang}
was born in Hebei, China. He received the B.E. degree and M.S.degree in Control Science and Engineering from Harbin Institute of Technology, China, in 2016 and 2018, respectively. He is currently studying for a Ph.D degree in Electronic science and technology in National Key laboratory of Tunable Laser Technology, Harbin Institute of Technology. His research interests include image restoration and enhancement, SLAM.
\end{IEEEbiography}

\begin{IEEEbiography}[{\includegraphics[width=1in,height=1.25in,clip,keepaspectratio]{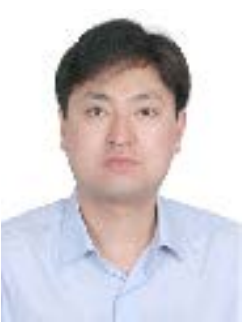}}]{Xiaoguang Di}
was born in Heilongjiang,China. He received the M.S. and Ph.D degree in navigation, guidance and control from Northwestern Polytechnical University, China, in 1999 and 2004, respectively. He is currently an associate professor with the Control and Simulation Center, Harbin Institute of Technology, Where he is in charge of courses in digital image processing and computer vision. His current research interests include real-time image restoration and enhancement, 3D object detection and recognition, SLAM. Prof. Di is a member of China Simulation Federation and Chinese Society of Astronautics.
\end{IEEEbiography}

\begin{IEEEbiography}[{\includegraphics[width=1in,height=1.25in,clip,keepaspectratio]{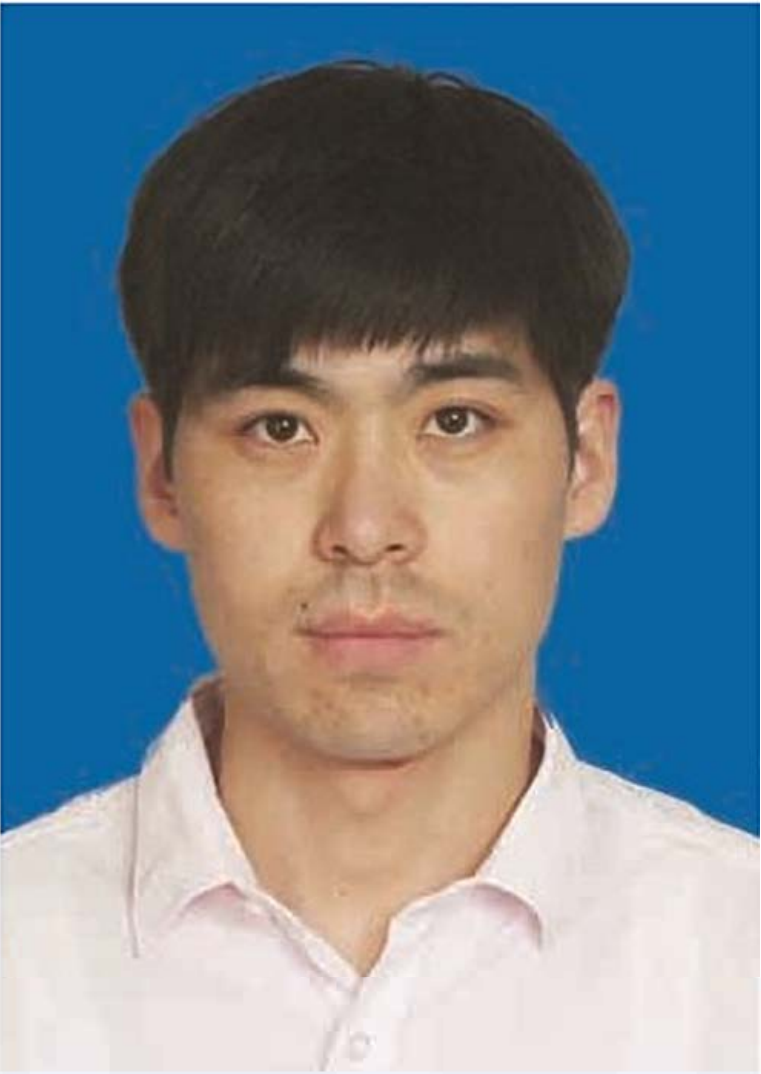}}]{Bin Zhang}
 was born in Gansu, China. He received the B.Sc. degree in apply physics and the M.Sc. degree in physics from China University of Petroleum, in 2010 and 2013, respectively. He is currently studying for a PhD degree in National Key Laboratory of Tunable Laser Technology, Harbin Institute of Technology. His research interests include laser imaging and laser transmission.
\end{IEEEbiography}


\begin{IEEEbiography}[{\includegraphics[width=1in,height=1.25in,clip,keepaspectratio]{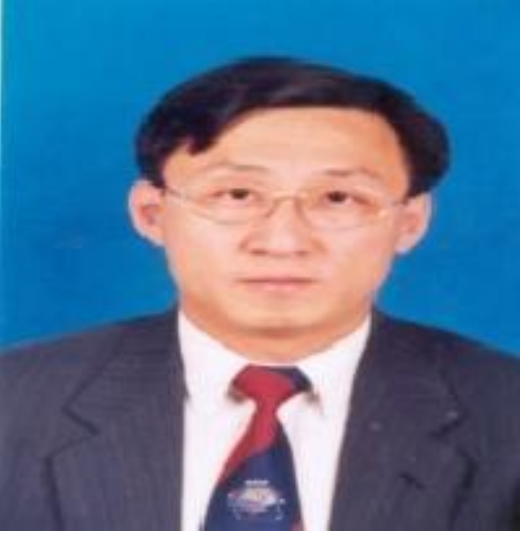}}]{Chunhui Wang}
Chunhui Wang received the BS, MS, and PhD degrees from Harbin Institute of Technology in 1987, 1991, and 2005，respectively. He is currently a processor and Deputy Director, Institute of Optoelectronic Technology, Harbin Institute of Technology. His current major research interests are laser remote sensing, lidar, laser detection and recognition.
\end{IEEEbiography}





\clearpage
\end{CJK*}

\end{document}